\documentclass[10pt,twocolumn,letterpaper]{article}
\PassOptionsToPackage{table}{xcolor}

\usepackage[pagenumbers]{cvpr}
\usepackage{amsmath,amssymb,bm}

\usepackage{multirow}
\usepackage{array}
\usepackage{arydshln}
\usepackage{algorithm}
\usepackage{algpseudocode}
\definecolor{bestgreen}{RGB}{166,219,160}
\definecolor{secondgreen}{RGB}{224,242,220}
\newcommand{\best}[1]{\cellcolor{bestgreen}{\textbf{#1}}}
\newcommand{\second}[1]{\cellcolor{secondgreen}{#1}}
\newcommand{\besttext}[1]{\colorbox{bestgreen}{\textbf{#1}}}
\newcommand{\secondtext}[1]{\colorbox{secondgreen}{#1}}

\definecolor{cvprblue}{rgb}{0.21,0.49,0.74}
\usepackage[breaklinks,colorlinks,allcolors=cvprblue]{hyperref}

\def\confName{CVPR}
\def\confYear{2027}

\title{SplatTether: Tightly Coupled Visual-Inertial Localization
with 3D Gaussian Splatting Maps via Learned Correspondence Uncertainty}

\author{
Taisei~Ando$^{1,2}$, Kenji~Koide$^{1}$, Taku~Okawara$^{1}$, Mizuho~Aoki$^{1}$,\\
Aoki~Takanose$^{1}$, Shuji~Oishi$^{1}$, Masashi~Yokozuka$^{1}$, and Atsushi~Yamashita$^{2}$\\[4pt]
$^{1}$National Institute of Advanced Industrial Science and Technology (AIST)\\
$^{2}$The University of Tokyo (UTokyo)\\[4pt]
\url{https://smrg-students.github.io/splattether-site/}
}

\makeatletter
\g@addto@macro\@maketitle{\begin{center}
  \includegraphics[width=\textwidth]{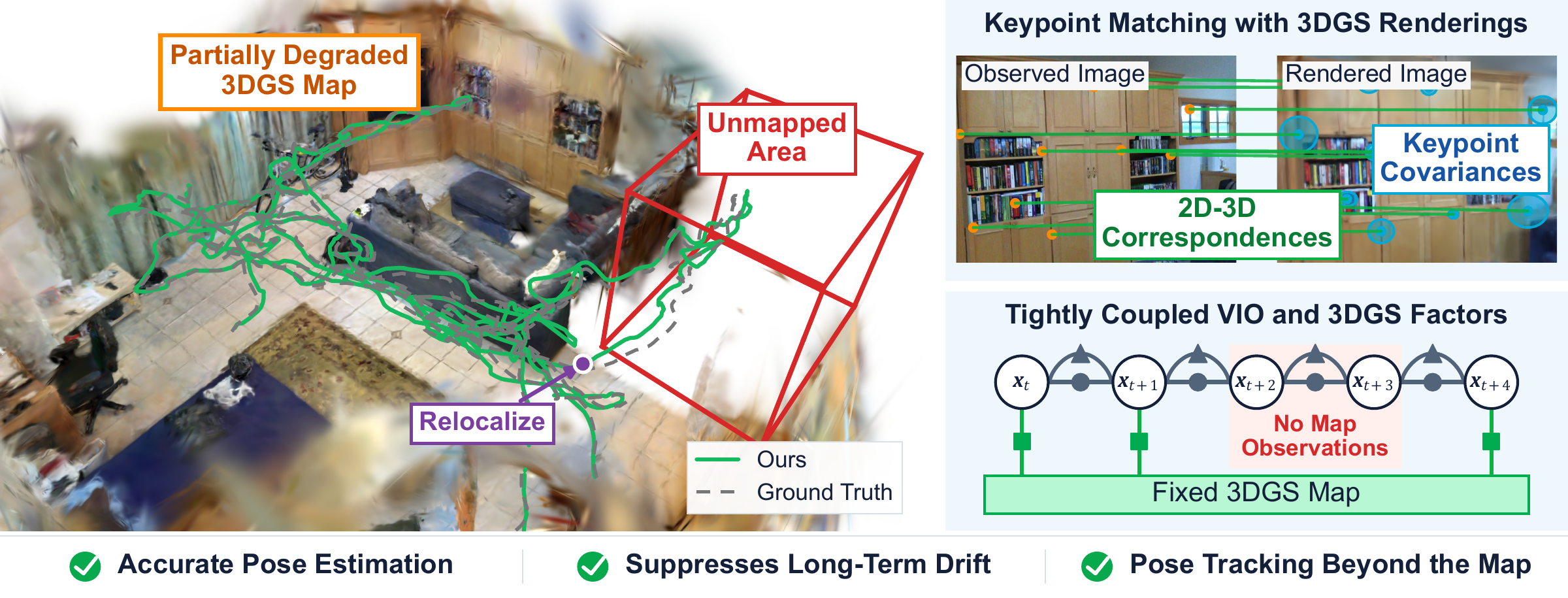}
  \captionsetup{hypcap=false}
  \captionof{figure}{SplatTether matches the observed image to an image rendered
    from a fixed 3DGS map and jointly optimizes states in a factor graph
    combining reprojection factors from
    \textcolor[RGB]{8,124,53}{2D-3D correspondences} with visual and inertial factors.
    It \textcolor[RGB]{0,120,160}{predicts positional covariances of keypoints
    on the rendered image} to model matching uncertainty and weight map factors.
    Even when the map contains \textcolor[RGB]{198,83,0}{degraded areas}, this
    formulation limits their influence, helping improve pose estimation
    accuracy and suppress long-term drift.
    SplatTether maintains pose estimation through VIO
    \textcolor[RGB]{214,40,40}{in unmapped areas} while using
    \textcolor[RGB]{123,63,160}{retrieval-based relocalization} for recovery when
    map matching becomes unreliable.}
  \label{fig:overview}
\end{center}
}
\makeatother

\begin{document}
\maketitle
\begin{abstract}
Map-based camera localization provides accurate pose estimates but is
vulnerable to erroneous or insufficient correspondences caused by map
inaccuracies or motion beyond the mapped area.
To address this problem, we propose SplatTether, which tightly couples
localization in a prebuilt 3D Gaussian Splatting (3DGS) map with local motion
estimation using visual-inertial odometry (VIO).
Our method jointly optimizes sliding-window states using a factor graph
that combines 2D-3D reprojection factors derived from the 3DGS map with
visual and inertial factors.
This tight coupling maintains pose estimation using VIO when map
correspondences are insufficient and suppresses drift using map constraints
when they are available.
To account for varying uncertainty across map correspondences, we train
a model to predict positional covariances for keypoints in rendered images.
We propagate these covariances to weight each reprojection factor according
to the uncertainty of the corresponding map observation.
Experiments demonstrate that integrating 3DGS map observations with VIO
reduces drift, while tight coupling and learned correspondence
uncertainty improve accuracy.

\end{abstract}
\section{Introduction}
\label{sec:introduction}

Visual localization estimates the camera pose within a known scene from
images and is essential for reliable navigation of autonomous mobile robots.
A common approach matches observed images against a prebuilt map, whose
fidelity to scene geometry and appearance directly affects localization
accuracy.

3D Gaussian Splatting (3DGS)~\cite{3dgs2023} enables fast, high-quality
novel-view rendering and has been used as a map representation for visual
localization~\cite{3dgsreloc2024,gsloc2024,splatloc2024,stdloc2025,gsvisloc2025,splathloc2026,3dgsuncertainty2026}.
However, degraded or incomplete 3DGS maps and environmental changes
can lead to insufficient or erroneous correspondences,
making map-based pose estimation unstable.
Maintaining map accuracy and coverage over time requires
additional image capture and updates in response to environmental changes.
Localization with reduced reliance on map quality and coverage is therefore
desirable when frequent map updates are impractical.

Combining map matching with local odometry~\cite{vinsmono2018,openvins2020,dmvio2022,supervins2025}
that operates independently of the map can maintain pose estimation when
map correspondences are insufficient.
A straightforward approach is to fuse map-based pose estimates with local
odometry.
However, aggregating map correspondences into a single pose observation prevents
individual reevaluation of their geometric constraints during optimization.
Correspondence uncertainty also varies across matches depending on map
quality, so uniform weighting cannot reflect these differences in state
estimation.
This motivates integrating map observations with odometry at the
correspondence level while accounting for the uncertainty of individual
correspondences.

We propose SplatTether, a visual-inertial localization method that tightly
couples 3DGS map-based localization with visual-inertial odometry
(VIO)~\cite{vinsmono2018}.
Figure~\ref{fig:overview} illustrates the concept.
Our method integrates map correspondences with visual and inertial
measurements in a common factor graph, suppressing drift through map
constraints while maintaining pose estimation through VIO when map
correspondences are insufficient.
It also weights each map correspondence by its predicted uncertainty to reduce
the influence of unreliable map observations.
When map matching based on jointly estimated poses becomes unreliable,
retrieval-based relocalization reestablishes the camera pose in the map,
allowing local map matching to resume.

Our main contributions are as follows:
\begin{itemize}
  \item \textbf{Tightly Coupled Map-Based Localization:}
  We combine 2D-3D reprojection factors based on 3DGS-rendered depth with
  visual and inertial factors in a common factor graph to jointly optimize
  sliding-window states.
  This incorporates the geometric constraints and uncertainty of individual
  map correspondences alongside visual and inertial constraints in state
  estimation.
  Map constraints thus suppress drift, while VIO maintains pose estimation
  when map correspondences are insufficient.

  \item \textbf{Learned Correspondence Uncertainty:}
  We predict keypoint covariance in 3DGS-rendered images from descriptors,
  descriptor differences, depth geometry, and matching confidence.
  We propagate this covariance through backprojection and reprojection
  to weight each correspondence factor according to the propagated
  reprojection covariance.

  \item \textbf{System Design with Relocalization:}
  After initialization, we render reference images at jointly estimated
  poses, providing nearby reference views for local map matching.
  If too few correspondences are available, we use image retrieval and PnP
  for relocalization.
  A validated pose is incorporated as a recovery pose factor, allowing
  local map matching to resume.
\end{itemize}

\begin{figure*}[t]
  \centering
  \includegraphics[width=.95\textwidth]{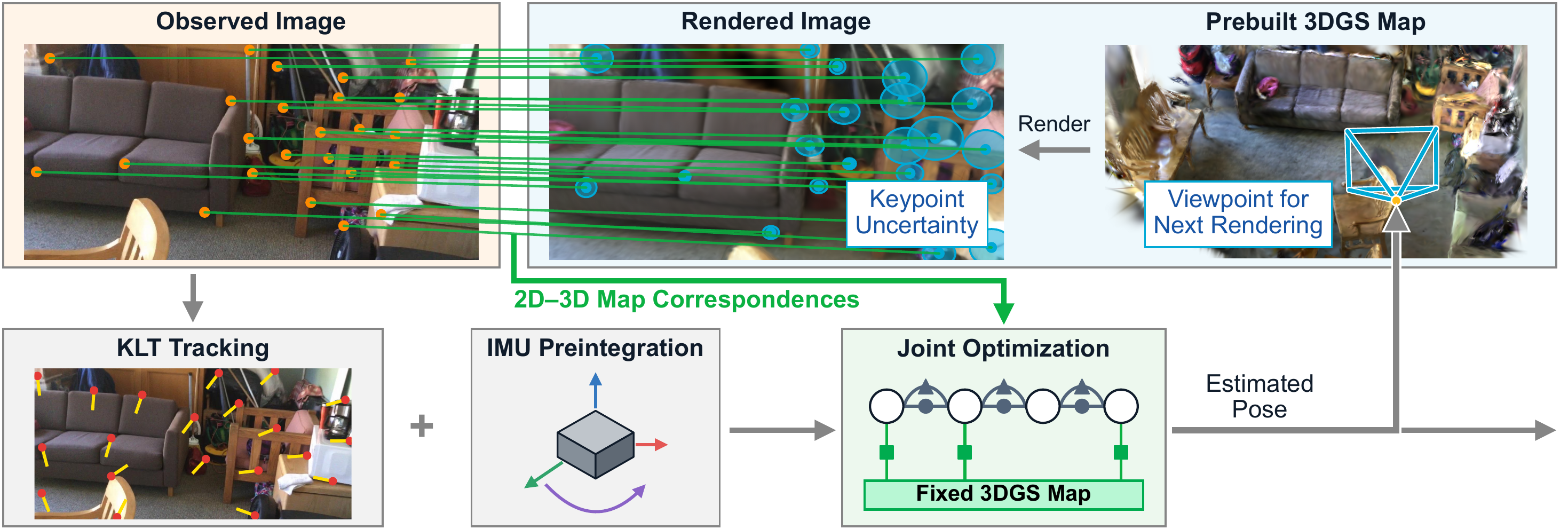}
  \caption{Overview of our method.
  The observed image is matched to a rendering of a prebuilt 3DGS map.
  Matched keypoints on the rendered image are backprojected using rendered
  depth to form 2D-3D correspondences.
  Joint optimization combines map factors weighted by predicted
  uncertainty (cyan ellipses) with KLT tracks and IMU preintegration.
  The estimated pose guides the next rendering.}
  \label{fig:method-overview}
\end{figure*}

\section{Related Work}

\subsection{Image-Based Map Localization}

Image-based map localization estimates camera poses using a prior map.
HLoc~\cite{hloc2019} combines image retrieval with local feature matching to estimate camera poses from 2D-3D correspondences with a structure-from-motion (SfM) map.
MASt3R~\cite{mast3r2024}, Reloc3r~\cite{reloc3r2025}, and VGGT~\cite{vggt2025} use learned models for image matching and pose estimation.
DSAC*~\cite{dsacstar2022} and ACE~\cite{ace2023} regress 3D scene coordinates from images, while iNeRF~\cite{inerf2021} directly aligns NeRF renderings with query images.

3DGS~\cite{3dgs2023} supports arbitrary-view rendering and serves as a map representation for localization.
3DGS-ReLoc~\cite{3dgsreloc2024} and GSLoc~\cite{gsloc2024} estimate camera poses through alignment with rendered images.
SplatLoc~\cite{splatloc2024}, STDLoc~\cite{stdloc2025}, GSVisLoc~\cite{gsvisloc2025}, and SplatHLoc~\cite{splathloc2026} use feature representations or synthesized views for pose estimation.
UGS-Loc~\cite{3dgsuncertainty2026} refines camera poses through pose hypothesis sampling and PnP-RANSAC guided by geometric uncertainty.

However, insufficient correspondences due to unmapped areas or environmental changes, and inaccurate matches due to poor map quality, destabilize pose estimation.

\subsection{Visual-Inertial Odometry}

VIO methods are broadly divided into filter-based and optimization-based approaches.
The filter-based MSCKF~\cite{msckf2007} uses feature observations across camera states, and OpenVINS~\cite{openvins2020} provides a VIO framework based on it.
The optimization-based OKVIS~\cite{okvis2015} and VINS-Mono~\cite{vinsmono2018} jointly optimize visual feature observations and IMU preintegration, while DM-VIO~\cite{dmvio2022} uses delayed marginalization.
SuperVINS~\cite{supervins2025} and MASt3R-Fusion~\cite{mast3rfusion2026} incorporate learned visual models into visual-inertial estimation.

However, VIO accumulates drift through local motion integration and cannot independently estimate poses in a prior map's frame.

\subsection{Integration of Map Localization and VIO}

One way to integrate map localization with VIO is to use poses estimated from the map in VIO state estimation.
The ROVIOLI frontend in maplab~\cite{maplab2018} fuses map-based pose estimates with visual-inertial odometry.
However, reducing map correspondences to pose observations prevents direct reevaluation of individual correspondence residuals and uncertainties as the VIO state changes.

Alternatively, Lynen et al.~\cite{vilocalization2020} incorporate global 2D-3D matches directly into EKF updates, while ORB-SLAM3~\cite{orbslam32021} jointly optimizes reprojection and inertial constraints during visual-inertial map merging.
For LiDAR maps, GLIL~\cite{glil2024} jointly optimizes scan-to-scan and scan-to-map registration factors with IMU factors to continuously estimate poses both within and beyond the mapped area.
Meanwhile, NeRF-VINS~\cite{nerfvins2024} tightly couples 2D-2D feature matches between real and nearby NeRF-rendered views with an MSCKF, accounting for rendering pose errors.
For keypoint uncertainty estimation, RaCo~\cite{raco2026} learns anisotropic keypoint covariances from homography-related image pairs.

In contrast, we constrain camera poses in the map frame with 2D-3D reprojection factors based on 3DGS-rendered depth and jointly optimize sliding window states with visual and inertial factors.
In this joint optimization, we propagate learned per-keypoint covariances
to reprojection residuals to adjust the strength and directional weighting
of each constraint according to observation uncertainty.

\section{Proposed Method}
\label{sec:method}

We jointly optimize sliding window states in a common factor graph combining
uncertainty-weighted 2D-3D correspondence factors from a 3DGS map with visual
and inertial factors. Figure~\ref{fig:method-overview} provides an overview of our method.

\subsection{State Representation}

Let $G$, $L$, $B$, and $C$ denote the 3DGS map, local VIO, IMU, and camera
coordinate frames, respectively. VIO first initializes its states in the local VIO frame $L$
using VINS-Mono~\cite{vinsmono2018}. Once the map frame initialization
in Section~\ref{subsec:map-frame-initialization} succeeds, a shared transform
${}^{G}\bm{T}_{L}$ maps the sliding window states into the map frame $G$, where they are
subsequently updated.

We index each state in the sliding window by $k$ and denote its observed
image by $\bm{I}_k^{\mathrm{obs}}$. Let $\bm{T}_k$ be the corresponding IMU pose in $G$
and ${}^{B}\bm{T}_{C}$ the camera to IMU extrinsic transform. The camera pose
is $\bm{T}_{C,k}=\bm{T}_k{}^{B}\bm{T}_{C}$.
We define the sensor state in the map frame as follows:
\begin{equation}
\bm{x}_k=
\left(
\bm{T}_k,
\bm{v}_k,
\bm{b}_k
\right).
\end{equation}
We denote the rotation and translation components of $\bm{T}_k$
by $\bm{R}_k$ and $\bm{p}_k$, respectively.
The velocity $\bm{v}_k$ is expressed in $G$, and $\bm{b}_k$ contains
the accelerometer and gyroscope biases.

\subsection{Correspondences from the 3DGS Map}
\label{subsec:correspondence-generation}

To obtain 2D-3D correspondences from the 3DGS map, we render a reference
image from a viewpoint close to that of the observed image. We build an image
retrieval database beforehand by rendering the 3DGS map from multiple viewpoints and
store a NetVLAD~\cite{netvlad2018} descriptor and camera pose for each image.
Before map frame initialization, we retrieve database images similar to the
observed image and use their poses as candidate viewpoints for rendering.
Using image retrieval for every subsequent frame, however, would require
repeated global searches and ignore the motion constraints between frames.
After initialization, we use the camera pose derived from the current state
estimate, $\bm{T}_{C,k}$, as the rendering viewpoint for local matching.
If correspondences are insufficient, we switch to image retrieval at the
next matching attempt.

Let $\bm{T}_k^{\mathrm{rend}}$ denote the rendering camera pose in $G$,
with rotation $\bm{R}_k^{\mathrm{rend}}$ and translation $\bm{t}_k^{\mathrm{rend}}$.
We render a reference image $\bm{I}_k^{\mathrm{rend}}$ from the
3DGS map, extract features from $\bm{I}_k^{\mathrm{obs}}$ and $\bm{I}_k^{\mathrm{rend}}$, and obtain
matches $(\bm{u}_{i,k}^{\mathrm{obs}},\bm{u}_{i,k}^{\mathrm{rend}},s_{i,k})$.
We extract features with XFeat~\cite{xfeat2024} and match them with
LighterGlue, a lightweight variant of LightGlue~\cite{lightglue2023}
trained for XFeat descriptors. The first two entries are keypoint coordinates
in the observed and reference images, and $s_{i,k}$ is the match confidence.

We lift each reference keypoint
$\bm{u}_{i,k}^{\mathrm{rend}}=[u_i^{\mathrm{rend}},v_i^{\mathrm{rend}}]^\top$ into 3D using the mean depth from
rendering. If $n_i$ Gaussians contribute to the pixel $\bm{u}_{i,k}^{\mathrm{rend}}$, with depths $d_j$ along
the rendering camera's optical axis and
compositing weights $w_j$, we compute the opacity-normalized mean depth as
follows:
\begin{equation}
\mu_i
=
\frac{\sum_{j=1}^{n_i}w_jd_j}{\sum_{j=1}^{n_i}w_j}.
\end{equation}
Using the rendering camera intrinsic matrix $\bm{K}^{\mathrm{rend}}$ and pose
$\bm{T}_k^{\mathrm{rend}}$, we backproject and transform keypoints in the
reference image to obtain map points:
\begin{equation}
{}^{G}\bm{p}_i
=
\bm{R}_k^{\mathrm{rend}}
\mu_i(\bm{K}^{\mathrm{rend}})^{-1}
\bigl[u_i^{\mathrm{rend}},\,v_i^{\mathrm{rend}},\,1\bigr]^{\top}
+\bm{t}_k^{\mathrm{rend}}.
\end{equation}
When using the current state estimate as the rendering viewpoint, we convert
matched keypoints in both images to bearing vectors using the processed image
intrinsics, check their angular consistency, and use only consistent matches
as map observations. If fewer than the minimum remain, we use image retrieval
to select the rendering viewpoint at the next matching attempt.
We apply RANSAC PnP to the 2D-3D correspondences from the retrieved viewpoint
and check the estimated pose using the inlier count, inlier ratio, and
reprojection error. We retain the first valid pose and repeat retrieval at
the next matching attempt. If the time interval is within its bound and
retrieval and VIO yield consistent relative motions, we add a recovery pose
factor only for the second pose and resume local matching.

\subsection{Covariance of Keypoints on Rendered Images}
\label{subsec:uncertainty-model}

Map coverage and rendering viewpoint affect keypoint location uncertainty,
which varies across correspondences. We therefore use a small multilayer
perceptron, termed the covariance head
$h^{\mathrm{rend}}$, to predict a covariance for each keypoint on the rendered
image. It takes the rendered image XFeat descriptor $\bm{d}_i^{\mathrm{rend}}$, its
elementwise absolute difference from the observed image descriptor
$|\bm{d}_i^{\mathrm{obs}}-\bm{d}_i^{\mathrm{rend}}|$, depth geometry $\bm{g}_i$, and LighterGlue match
confidence $s_{i,k}$ and outputs three scalars $z_{i,1},z_{i,2},z_{i,3}$.
The depth geometry comprises the mean and variance of rendered depth, the
variance of dominant depth, depth ambiguity, and the relative difference
between mean and dominant depth.

From these outputs, we construct the covariance of a keypoint on the rendered image
$\bm{\Sigma}_i^{\mathrm{rend}}\in\mathbb{R}^{2\times2}$ as follows:
\begin{align}
\bm{L}_i
&=\begin{bmatrix}
\log(1+e^{z_{i,1}}) & 0\\
z_{i,2} & \log(1+e^{z_{i,3}})
\end{bmatrix},\\
\bm{\Sigma}_i^{\mathrm{rend}}
&=\bm{L}_i\bm{L}_i^{\top}+\delta_{\mathrm{cov}}\bm{I}_2.
\end{align}
Here, $\delta_{\mathrm{cov}}>0$ is a numerical floor.

Let $\bm{J}_i^{\mathrm{bp}}\in\mathbb{R}^{3\times2}$ be the backprojection Jacobian
with respect to image coordinates at fixed mean depth $\mu_i$.
The map point covariance ${}^{G}\bm{\Sigma}_{\mathrm{p},i}\in\mathbb{R}^{3\times3}$ is:
\begin{equation}
{}^{G}\bm{\Sigma}_{\mathrm{p},i}
=
\bm{R}_k^{\mathrm{rend}}
\bm{J}_i^{\mathrm{bp}}
\bm{\Sigma}_i^{\mathrm{rend}}
\left(\bm{J}_i^{\mathrm{bp}}\right)^{\top}
\left(\bm{R}_k^{\mathrm{rend}}\right)^{\top}.
\end{equation}
Through this propagation, the head enables reprojection factors to weight
each correspondence by the magnitude and direction of its keypoint location
uncertainty.

\subsection{Joint Optimization}
\label{subsec:correspondence-factor}

Our method takes into account map, visual, and inertial observations in a common
factor graph and jointly optimizes the sliding window states.

For local matching, we transform a map point ${}^{G}\bm{p}_i$ into the camera frame as
${}^{C}\bm{p}_{i,k}=\left(\bm{T}_{C,k}\right)^{-1}{}^{G}\bm{p}_i$.
Using the projection function $\pi(\cdot)$, we define the reprojection
residual as follows:
\begin{equation}
\bm{r}_{i,k}
=
\pi\left({}^{C}\bm{p}_{i,k}\right)
-\bm{u}_{i,k}^{\mathrm{obs}}.
\end{equation}
We account for map point uncertainty in the residual covariance $\bm{S}_{i,k}$. Let
$\bm{J}_{\pi,i}$ be the projection Jacobian at ${}^{C}\bm{p}_{i,k}$.
Projecting the map point covariance into the image plane gives:
\begin{equation}
\bm{S}_{i,k}
=
\bm{J}_{\pi,i}
\left(\bm{R}_{C,k}\right)^{\top}
{}^{G}\bm{\Sigma}_{\mathrm{p},i}
\bm{R}_{C,k}
\bm{J}_{\pi,i}^{\top}.
\end{equation}
Because the projection depends on the current camera pose, we recompute
$\bm{S}_{i,k}$ at each pose evaluated during optimization and apply a numerical
eigenvalue floor $\epsilon>0$.
During linearization, we hold the covariance fixed and omit derivatives
of its pose dependence.

We define the per-correspondence cost using the squared Mahalanobis distance:
\begin{equation}
e_{i,k}
=
\frac{1}{2}
\bm{r}_{i,k}^{\top}\widetilde{\bm{S}}_{i,k}^{-1}\bm{r}_{i,k}.
\end{equation}
Here, $\widetilde{\bm{S}}_{i,k}$ is $\bm{S}_{i,k}$ with eigenvalues floored
at $\epsilon$.

Large pose errors can prevent recovery through local correspondence optimization.
We transform the accepted camera pose $\hat{\bm{T}}_{C,k}$ into the IMU pose
$\hat{\bm{T}}_k=\hat{\bm{T}}_{C,k}({}^{B}\bm{T}_{C})^{-1}$,
with rotation $\hat{\bm{R}}_k$ and translation $\hat{\bm{p}}_k$.
We define the residual of the recovery pose factor as follows:
\begin{equation}
\bm{r}_{k}^{\mathrm{pose}}=
\begin{bmatrix}
\hat{\bm{R}}_k^{\top}
\left(\bm{p}_k-\hat{\bm{p}}_k\right)\\
\operatorname{Log}\!\left(\hat{\bm{R}}_k^{\top}\bm{R}_k\right)
\end{bmatrix}.
\end{equation}
We weight this residual using a pose information matrix
$\bm{\Lambda}_{k}^{\mathrm{pose}}$ derived from the accepted
map correspondences and their propagated covariances.
The information matrix is transformed to the IMU pose residual
coordinates and capped to prevent excessive confidence.
We define the recovery pose cost as follows:
\begin{equation}
e_k^{\mathrm{pose}}
=
\frac{1}{2}
\left(\bm{r}_k^{\mathrm{pose}}\right)^{\top}
\bm{\Lambda}_k^{\mathrm{pose}}
\bm{r}_k^{\mathrm{pose}}.
\end{equation}
For each accepted map observation, we register correspondence factors for
local matching or a pose factor for retrieval-based recovery, avoiding
duplicate use of the same observation.

Together with the visual and inertial factors, these selected map factors
define the objective over the set of sliding-window states
$\mathcal{X}$, with map points fixed:
\begin{equation}
\mathcal{X}^{*}
=\operatorname*{arg\,min}_{\mathcal{X}}
\left[
\begin{aligned}
&E_{\mathrm{vio}}(\mathcal{X})
+\sum_{(i,k)\in\mathcal{C}} e_{i,k}(\mathcal{X})
\\
&+\sum_{k\in\mathcal{R}} e_k^{\mathrm{pose}}(\mathcal{X})
\end{aligned}
\right].
\end{equation}
Here, $\mathcal{C}$ is the set of selected map correspondences,
$\mathcal{R}$ is the set of states with active recovery pose factors, and
$E_{\mathrm{vio}}$ is the total visual, inertial, and marginalization cost.
Local visual feature inverse depths are jointly estimated as auxiliary variables.

\subsection{Initial Pose Estimation}
\label{subsec:map-frame-initialization}

Before joint optimization with map correspondence factors, we estimate the transform
${}^{G}\bm{T}_{L}$ from the local frame $L$ to the map frame $G$.
We first render reference images from viewpoints selected by image retrieval
(Section~\ref{subsec:correspondence-generation}), match them to observed images,
and collect 2D-3D correspondences across multiple frames.
We then obtain an initial estimate of ${}^{G}\bm{T}_{L}$ from the retrieved
camera poses and corresponding VIO poses, and refine it using the reprojection
errors of the collected correspondences.

Inspired by GLIL~\cite{glil2024}, we use VIO-initialized gravity and the known
map-frame gravity direction to fix roll and pitch. This leaves four degrees of
freedom, namely translation ${}^{G}\bm{t}_{L}$ and yaw $\psi$.
To reduce the effect of mismatches in a single frame, we search for
hypotheses across multiple frames and minimize a covariance-weighted
reprojection cost $e_{i,k}^{\mathrm{init}}$ over the reprojection consensus set
$\mathcal{C}_k^{\mathrm{inlier}}$:
\begin{equation}
\left({}^{G}\bm{t}_{L}^{*},\psi^{*}\right)
=
\operatorname*{arg\,min}_{{}^{G}\bm{t}_{L},\psi}
\sum_{k\in\mathcal{I}}
\sum_{i\in\mathcal{C}_k^{\mathrm{inlier}}}
e_{i,k}^{\mathrm{init}}\!\left({}^{G}\bm{T}_{L}\right).
\end{equation}
Here, $\mathcal{I}$ is the set of frames used for initialization.
The initialization cost has the same form as that in
Section~\ref{subsec:correspondence-factor}. We validate the alignment
before registering the initial map factors. If
registration succeeds, we finalize
${}^{G}\bm{T}_{L}$ and transform the VIO states into the map frame.
If either IMU bias norm exceeds its threshold, we reset the estimator and
repeat VIO and map frame initialization.

\subsection{Self-supervised Training of the Covariance Head}
\label{subsec:uncertainty-training}

Regions insufficiently observed during 3DGS map construction can appear
blurred or have ambiguous boundaries when rendered from new viewpoints,
which can reduce keypoint localization accuracy. Treating these effects as ordinary
image blur alone may miss geometric uncertainty in weakly constrained map
Gaussians and underestimate keypoint location covariance. We therefore train
$\bm{\Sigma}_i^{\mathrm{rend}}$ using matches between real and rendered images.

\begin{figure}[t]
  \centering
  \includegraphics[width=.9\linewidth]{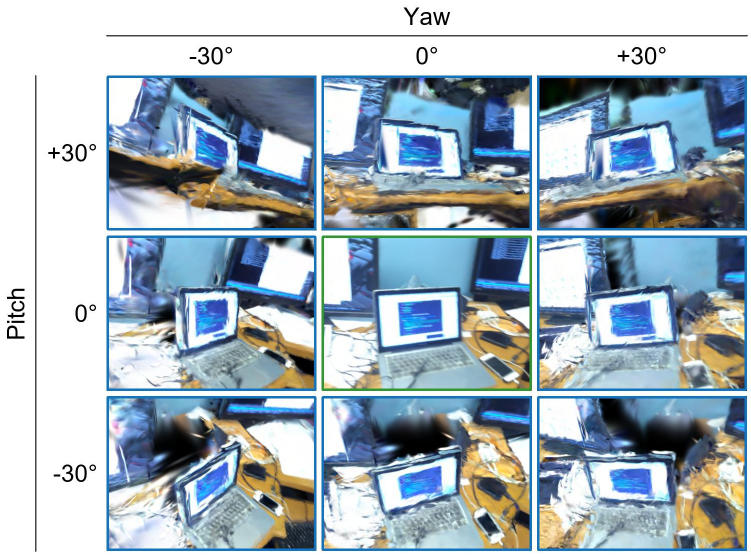}
  \caption{Nine rendered views at yaw and pitch offsets of $-30^\circ$, $0^\circ$, and $+30^\circ$. The central view uses the observed camera pose.}
  \label{fig:rendering-covariance-training}
\end{figure}

We generate virtual viewpoints from RGB-D frames excluded from map
construction and render 3DGS images at their known poses and nearby views.
Figure~\ref{fig:rendering-covariance-training} shows a relatively sharp
central view and blur and ambiguous boundaries in displaced views.

We backproject real image keypoints using sensor depth and project them into
each virtual view to obtain $\bm{u}_i^{\mathrm{gt}}$. Let $\bm{u}_i^{\mathrm{rend}}$ be
the matched position in the rendered image. The pixel residual is
$\bm{e}_i^{\mathrm{u}}=\bm{u}_i^{\mathrm{rend}}-\bm{u}_i^{\mathrm{gt}}$.
We retain matches consistent with the ground truth pose and sensor depth.

We train this head over the set of valid matches $\mathcal{M}$ by minimizing
the loss $\mathcal{L}$, defined as the Gaussian negative log likelihood of
their pixel residuals:
\begin{align}
\mathcal{L}
&=\frac{1}{|\mathcal{M}|}\sum_{i\in\mathcal{M}}\ell_i,\\
\ell_i
&=\frac{1}{2}(\bm{e}_i^{\mathrm{u}})^\top
(\bm{\Sigma}_i^{\mathrm{rend}})^{-1}\bm{e}_i^{\mathrm{u}}
+\frac{1}{2}\log\det\bm{\Sigma}_i^{\mathrm{rend}}
+\log(2\pi).
\end{align}

\begin{figure*}[t]
  \centering
  \includegraphics[width=.75\textwidth]{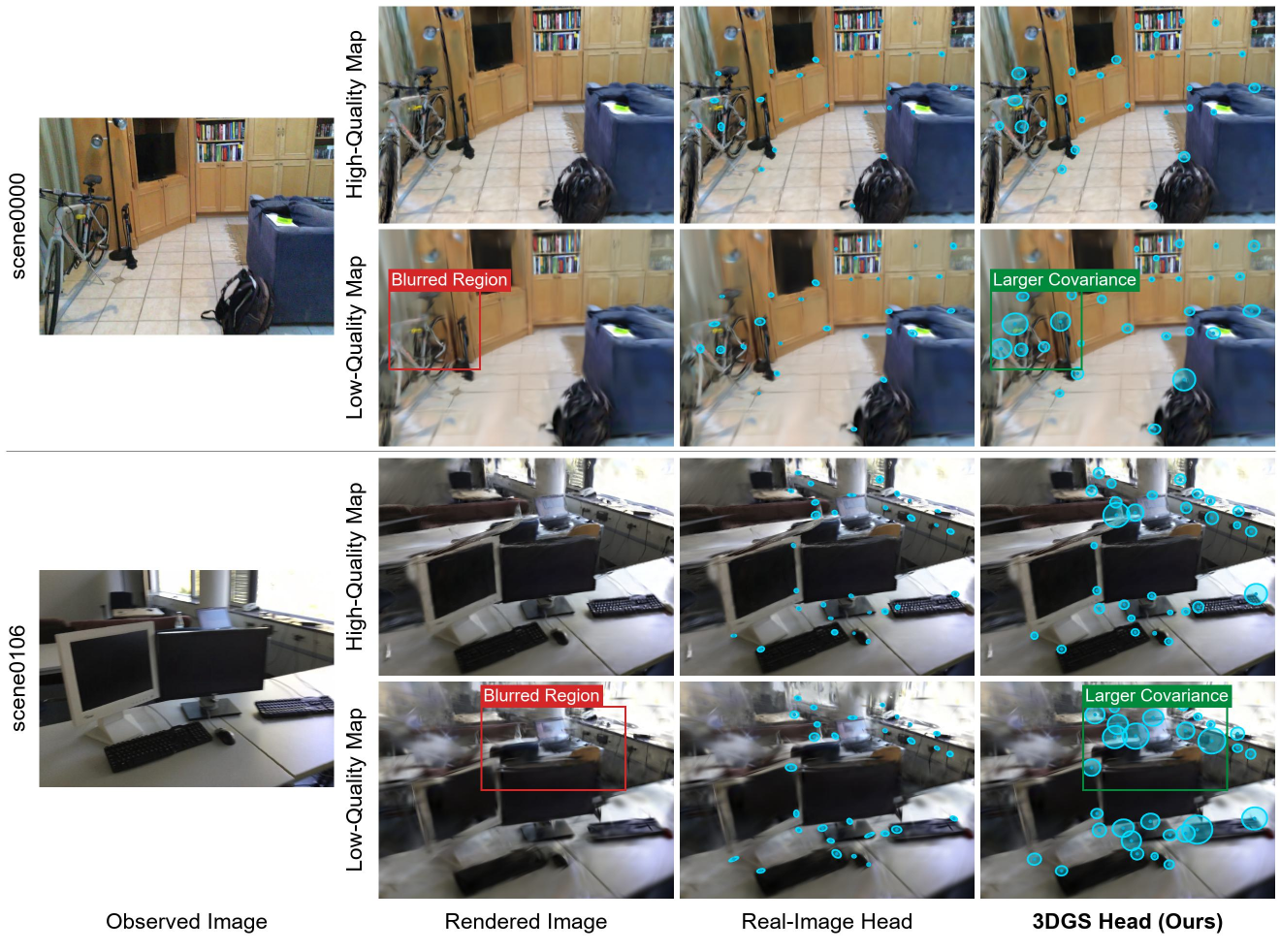}
  \caption{Covariance head comparison on renderings from high-quality and
  low-quality 3DGS maps of ScanNet scenes 0000 and 0106. Cyan ellipses indicate
  the 95\% probability regions implied by the predicted covariances of
  keypoints on the rendered images.}
  \label{fig:qualitative-covariance}
\end{figure*}

\section{Experiments}
\label{sec:experiment}

We evaluate the proposed method against other localization methods and
examine the contributions of its components through ablation experiments.

\subsection{Experimental Setup}

\medskip
\noindent\textbf{Datasets.}
We evaluate on an in-house dataset with recorded IMU measurements and on
the public ScanNet dataset~\cite{scannet2017} with synthetic IMU measurements.
The evaluation scenes are excluded from covariance head training and
model selection (Section~\ref{subsec:uncertainty-training}).

We collected the in-house dataset in a laboratory (Room), a corridor
(Corridor), and an outdoor parking lot (Parking).
For each environment, we prepared a mapping segment and two evaluation
sequences, Easy and Hard. Hard includes more challenging segments with
prolonged occlusions or large camera rotations.
We recorded the data using a sensor rig equipped with a Livox Mid-360 and
an Intel RealSense D455 and processed the images at $640\times320$ pixels.
We calibrated the LiDAR-to-camera
extrinsics using separate calibration data and the LiDAR-camera calibration
toolbox~\cite{calib2023}.

For ScanNet, we assign different capture sessions to map construction and
evaluation and synthesize IMU data from camera trajectories.
To remove black borders caused by undistortion, we resize ScanNet RGB images
to $640\times480$ pixels and crop 16 pixels from each edge, yielding
$608\times448$ pixels. We adjust the camera intrinsics accordingly.

\medskip
\noindent\textbf{Experimental Conditions.}
Experiments are conducted on a laptop with an AMD Ryzen 9 7940HS CPU and an
NVIDIA GeForce RTX 4080 Laptop GPU with 12 GB of VRAM.
We construct the 3DGS maps for our method offline using a mapping pipeline
based on DN-Splatter~\cite{dnsplatter2025}.

For localization, we compare methods based on reference images or scene
coordinates~\cite{hloc2019,ace2023,mast3r2024,reloc3r2025,vggt2025}
and 3DGS maps~\cite{splatloc2024,stdloc2025}.
We compare with geometric VIO and visual-inertial SLAM
baselines~\cite{vinsmono2018,orbslam32021,openvins2020,dmvio2022}
and visual-inertial methods using learned visual
models~\cite{supervins2025,mast3rfusion2026}.
For each dataset, we use common IMU noise coefficients across all
visual-inertial methods, including ours, and convert them to each
implementation's parameter definitions.

\medskip
\noindent\textbf{Evaluation Protocol.}
To evaluate continuous pose estimation, we compare our method with
map-based localization and visual-inertial pose estimation methods using
the RMSE of the translational absolute trajectory error (ATE).
For localization, we evaluate pose estimates directly in the map frame
and exclude those at least $100\,\mathrm{m}$ from the mapped area as outliers.
For visual-inertial pose estimation, we align each trajectory, including
ours, to ground truth using a 4-DoF transform comprising translation
and rotation about gravity, determined from all matched positions
along the trajectory~\cite{vioeval2018}.
We use ``-'' for runs with coverage below 80\% or an ATE of at least
$100\,\mathrm{m}$.
We report the median of three runs, ranking failures last.
The average is also marked as ``-'' if any scene has ``-''.
In the result tables, the \besttext{best} values are bold with a darker green
background, and the \secondtext{second-best} values have a lighter green
background.

\subsection{Qualitative Results}

Figure~\ref{fig:qualitative-covariance} compares our 3DGS head with a head
trained on pairs synthesized from real images (\textit{Real-image head} in
Table~\ref{tab:gsvins_ablation_results}), using renderings from high-quality
and low-quality maps of two ScanNet scenes.
In strongly blurred regions of the low-quality renderings, the real-image
head predicts small covariances, whereas ours predicts large covariances
that reduce correspondence weights and limit the influence of blurred
regions on optimization.

We also evaluate pose continuity in scene 0000 with a deliberately
unmapped region (Figure~\ref{fig:overview}).
Map-only localization baselines had gaps in their pose outputs or
localized to incorrect map regions, while our method maintained pose
estimation through VIO with an ATE of 0.055~m.

\subsection{Quantitative Evaluation}

\begin{table}[t]
  \centering
  \caption{Translation ATE RMSE [m]$\downarrow$ on in-house data in the map coordinate frame
  (median of three runs).}
  \label{tab:comparison_inhouse_localization}
  \footnotesize
  \setlength{\tabcolsep}{2pt}
  \begin{tabular}{@{}lccccccc}
    \toprule
    & \multicolumn{2}{c}{Room} & \multicolumn{2}{c}{Corridor} & \multicolumn{2}{c}{Parking} & \multirow{2}{*}{Avg.} \\
    \cmidrule(lr){2-3}\cmidrule(lr){4-5}\cmidrule(lr){6-7}
    Method & Easy & Hard & Easy & Hard & Easy & Hard & \\
    \midrule[0.3pt]
    HLoc & 3.776 & 6.474 & 3.308 & - & 4.886 & 6.645 & - \\
    ACE & 2.880 & 4.235 & 11.922 & 30.985 & 15.396 & 22.745 & 14.694 \\
    MASt3R loc. & \second{0.216} & \second{0.268} & \second{0.992} & - & \second{1.087} & \second{1.412} & - \\
    Reloc3r & 1.375 & 2.481 & 3.088 & \second{5.434} & 6.101 & 8.625 & \second{4.517} \\
    VGGT loc. & 1.038 & 2.152 & - & - & - & - & - \\
    \addlinespace[2pt]
    \cdashline{1-8}[1.2pt/1.8pt]
    \addlinespace[2pt]
    SplatLoc & 1.602 & - & 3.453 & - & - & - & - \\
    STDLoc & 4.964 & 9.476 & 3.037 & - & 18.267 & 21.810 & - \\
    \textbf{Ours} & \best{0.075} & \best{0.090} & \best{0.219} & \best{0.896} & \best{0.289} & \best{0.748} & \best{0.386} \\
    \bottomrule
  \end{tabular}
\end{table}

\begin{table}[t]
  \centering
  \caption{Translation ATE RMSE [m]$\downarrow$ on ScanNet in the map coordinate frame
  (median of three runs).}
  \label{tab:comparison_scannet_localization}
  \footnotesize
  \setlength{\tabcolsep}{2pt}
  \begin{tabular}{@{}lccccccc}
    \toprule
    Method & 0000 & 0059 & 0106 & 0169 & 0181 & 0207 & Avg. \\
    \midrule[0.3pt]
    HLoc & 0.278 & 0.326 & 0.608 & 1.181 & 0.278 & 0.265 & 0.489 \\
    ACE & 1.390 & 0.243 & 1.086 & 1.383 & 1.168 & 0.177 & 0.908 \\
    MASt3R loc. & 0.120 & 0.161 & 0.343 & \second{0.120} & 0.321 & \second{0.113} & \second{0.196} \\
    Reloc3r & 0.278 & 0.197 & 0.429 & 0.846 & 0.598 & 0.243 & 0.432 \\
    VGGT loc. & \second{0.068} & 0.223 & \second{0.288} & 0.830 & \second{0.175} & 0.141 & 0.288 \\
    \addlinespace[2pt]
    \cdashline{1-8}[1.2pt/1.8pt]
    \addlinespace[2pt]
    SplatLoc & 0.089 & 0.241 & - & 0.364 & 0.413 & 0.146 & - \\
    STDLoc & 2.805 & \second{0.146} & 7.351 & 2.116 & 4.401 & 1.909 & 3.121 \\
    \textbf{Ours} & \best{0.055} & \best{0.126} & \best{0.159} & \best{0.117} & \best{0.093} & \best{0.110} & \best{0.110} \\
    \bottomrule
  \end{tabular}
\end{table}

\medskip
\noindent\textbf{Localization.}
As shown in Table~\ref{tab:comparison_inhouse_localization}, our method
achieves the lowest ATE in all six conditions on the in-house dataset.
Some baselines achieve the second-best accuracy on Easy sequences with
relatively little occlusion, but their errors increase on Hard sequences
with occlusions or brief views of the sky or ceiling.
Despite larger errors on Hard sequences than on Easy sequences, our
method maintains lower ATE than the baselines in all conditions.

As shown in Table~\ref{tab:comparison_scannet_localization}, our method
achieves the lowest ATE in all six ScanNet scenes.
For scene 0000 in the qualitative evaluation, removing part of the map
yielded an ATE comparable to that under the standard condition.
This suggests that VIO can maintain trajectory accuracy despite
insufficient map correspondences in segments where motion can be
estimated reliably.
Together with the in-house results, this indicates that combining map
observations with visual-inertial estimation is particularly beneficial
under occlusions or large camera rotations.

\begin{table}[t]
  \centering
  \caption{Translation ATE RMSE [m]$\downarrow$ on in-house data after 4-DoF alignment
  (median of three runs).}
  \label{tab:comparison_inhouse_sequential}
  \footnotesize
  \setlength{\tabcolsep}{2pt}
  \begin{tabular}{@{}lccccccc}
    \toprule
    & \multicolumn{2}{c}{Room} & \multicolumn{2}{c}{Corridor} & \multicolumn{2}{c}{Parking} & \multirow{2}{*}{Avg.} \\
    \cmidrule(lr){2-3}\cmidrule(lr){4-5}\cmidrule(lr){6-7}
    Method & Easy & Hard & Easy & Hard & Easy & Hard & \\
    \midrule[0.3pt]
    VINS-Mono & 0.137 & 0.275 & \second{0.932} & 67.476 & \second{0.659} & 33.865 & 17.224 \\
    OpenVINS & \second{0.110} & \second{0.203} & - & - & - & - & - \\
    ORB-SLAM3 & - & - & - & - & - & - & - \\
    DM-VIO & 0.350 & - & 7.364 & - & - & \second{1.517} & - \\
    \addlinespace[2pt]
    \cdashline{1-8}[1.2pt/1.8pt]
    \addlinespace[2pt]
    SuperVINS & 0.256 & - & - & - & - & 40.659 & - \\
    MASt3R-Fusion & 0.525 & 0.488 & 0.950 & \second{1.072} & 1.243 & 2.337 & \second{1.103} \\
    \textbf{Ours} & \best{0.067} & \best{0.087} & \best{0.209} & \best{0.837} & \best{0.244} & \best{0.670} & \best{0.352} \\
    \bottomrule
  \end{tabular}
\end{table}

\begin{table}[t]
  \centering
  \caption{Translation ATE RMSE [m]$\downarrow$ on ScanNet after 4-DoF alignment
  (median of three runs).}
  \label{tab:comparison_scannet_sequential}
  \footnotesize
  \setlength{\tabcolsep}{2pt}
  \begin{tabular}{@{}lccccccc}
    \toprule
    Method & 0000 & 0059 & 0106 & 0169 & 0181 & 0207 & Avg. \\
    \midrule[0.3pt]
    VINS-Mono & 0.176 & 0.215 & \best{0.097} & \second{0.221} & \second{0.137} & \second{0.135} & \second{0.164} \\
    OpenVINS & \second{0.140} & \second{0.203} & - & 0.345 & - & - & - \\
    ORB-SLAM3 & - & - & - & - & - & - & - \\
    DM-VIO & - & - & 1.353 & - & 0.470 & - & - \\
    \addlinespace[2pt]
    \cdashline{1-8}[1.2pt/1.8pt]
    \addlinespace[2pt]
    SuperVINS & - & 7.454 & 11.854 & 0.385 & 0.434 & 0.235 & - \\
    MASt3R-Fusion & 0.257 & 0.240 & 0.333 & 0.241 & 0.139 & 0.187 & 0.233 \\
    \textbf{Ours} & \best{0.051} & \best{0.090} & \second{0.106} & \best{0.104} & \best{0.083} & \best{0.086} & \best{0.087} \\
    \bottomrule
  \end{tabular}
\end{table}

\medskip
\noindent\textbf{Visual-Inertial Pose Estimation.}
As shown in Table~\ref{tab:comparison_inhouse_sequential}, our method
achieves the lowest ATE in all six conditions on the in-house dataset.
VINS-Mono and MASt3R-Fusion also produce valid results in all six conditions,
but have larger errors on the Hard sequences in Corridor and Parking than
in Room.
Additionally, ORB-SLAM3 did not meet the coverage criterion on either dataset
due to delayed initialization.
Our method also incurs larger errors on Hard sequences than on Easy
sequences, but maintains lower ATE than the baselines.

As shown in Table~\ref{tab:comparison_scannet_sequential}, our method
achieves the lowest ATE in five of six ScanNet scenes.
However, VINS-Mono outperforms ours in scene 0106, possibly because
visual-inertial estimation alone already yields a small error, leaving
limited room for improvement from map observations.
These results suggest that constraining the trajectory with map observations
during estimation helps reduce visual-inertial errors, particularly in
challenging sequences with occlusions or large camera rotations.

\medskip
\noindent\textbf{Runtime and Memory Usage.}
\begin{table}[t]
  \centering
  \caption{Standalone localization runtime (including image I/O) and peak GPU memory.}
  \label{tab:runtime-memory-localization-scannet}
  \footnotesize
  \begin{tabular}{@{}lrr}
    \toprule
    Method & Time [ms/image]$\downarrow$ & Peak GPU [MiB]$\downarrow$ \\
    \midrule[0.3pt]
    HLoc & 175.3 & \second{1,362} \\
    ACE & \best{19.5} & \best{360} \\
    MASt3R loc. & 2,338.5 & 5,176 \\
    Reloc3r & 1,498.8 & 2,940 \\
    VGGT loc. & 451.9 & 7,226 \\
    SplatLoc & 231.6 & 2,708 \\
    STDLoc & 566.4 & 3,020 \\
    \textbf{Ours (Map only)} & \second{45.1} & 1,602 \\
    \bottomrule
  \end{tabular}
\end{table}

\begin{table}[t]
  \centering
  \caption{Processing times (mean and 95th percentile) and nominal execution
  rates of the proposed system. Frontend time includes image I/O.}
  \label{tab:runtime-memory-sequential-scannet}
  \footnotesize
  \setlength{\tabcolsep}{3pt}
  \begin{tabular}{@{}lrrr}
    \toprule
    Component & Rate [Hz] & Mean [ms] & P95 [ms] \\
    \midrule[0.3pt]
    VIO frontend & 30 & 9.7 & 14.5 \\
    VIO backend & 10 & 9.8 & 10.4 \\
    Map matching & 10 & 29.8 & 48.3 \\
    \bottomrule
  \end{tabular}
\end{table}

Table~\ref{tab:runtime-memory-localization-scannet} compares standalone
localization runtime and peak GPU memory on ScanNet scene 0059.
Our method averages $45.1\,\mathrm{ms}$ per image, faster than all baselines
except ACE. This short runtime may
reflect selecting a single rendering viewpoint through image
retrieval, followed by matching with XFeat and
LighterGlue. Its peak GPU memory is about 60\% of SplatLoc's and about
half of STDLoc's.

Table~\ref{tab:runtime-memory-sequential-scannet} reports the mean and
95th percentile processing times and nominal execution rates of our
integrated system with pose feedback for local matching on the same scene.
The 95th percentile processing time of each component is below its
corresponding execution interval.
These results support the feasibility of real-time estimation with the
proposed combination of map matching and VIO.

\medskip
\noindent\textbf{Ablation.}
To evaluate the contribution of each component, we compare our method with
the following variants:
\begin{itemize}
  \item \textbf{VIO only:} Uses no map factors.
  \item \textbf{Map only:} Localizes each image independently without VIO.
  \item \textbf{Loosely coupled:} Retains pose feedback but replaces
  correspondence factors with $\mathrm{SE}(3)$ pose factors derived
  from PnP-RANSAC estimates.
  \item \textbf{w/o pose feedback:} Uses retrieval without feeding the
  jointly optimized pose back to map matching.
  \item \textbf{w/o recovery:} Omits retrieval-based recovery.
  \item \textbf{Fixed cov.:} Replaces the predicted covariance of each
  keypoint on the rendered image with a fixed isotropic covariance.
  We use the VINS-Mono default standard deviation of $1.5\,\mathrm{px}$
  for default. For train, we aggregate the pixel residuals of valid training
  matches and use a shared standard deviation derived from their mean
  squared error.
  \item \textbf{Real-image head:} Replaces the covariance head with a
  real-image head trained on pairs generated from COCO~2017~\cite{coco2014}
  images using known homographies, with supervision inspired by
  RaCo~\cite{raco2026}.
\end{itemize}
Since \textit{VIO only} lacks a transform to the map frame, we compare all
methods using ATE after 4-DoF alignment.

\begin{table}[t]
  \centering
  \caption{Mean translation ATE RMSE [m]$\downarrow$ across six conditions per dataset
  after 4-DoF alignment (median of three runs per condition).
  Parentheses indicate valid conditions out of six.}
  \label{tab:gsvins_ablation_results}
  \footnotesize
  \setlength{\tabcolsep}{4pt}
  \begin{tabular}{@{}lcc}
    \toprule
    Method & In-house & ScanNet \\
    \midrule[0.3pt]
    VIO only & \second{1.096} & 0.168 \\
    Map only & - (1/6) & 0.516 \\
    Loosely coupled & - (4/6) & 0.213 \\
    w/o pose feedback & - (4/6) & - (5/6) \\
    w/o recovery & 1.163 & \second{0.088} \\
    \addlinespace[2pt]
    \cdashline{1-3}[1.2pt/1.8pt]
    \addlinespace[2pt]
    Fixed cov. (default) & 3.100 & 0.098 \\
    Fixed cov. (train) & 1.157 & 0.089 \\
    Real-image head & - (5/6) & 0.091 \\
    \textbf{Ours (full)} & \best{0.352} & \best{0.087} \\
    \bottomrule
  \end{tabular}
\end{table}

As shown in Table~\ref{tab:gsvins_ablation_results}, our method achieves
the lowest average ATE on both datasets.
In contrast, \textit{Map only} fails to meet the evaluation criteria in five in-house
conditions, suggesting that combining VIO with map matching supports
continuous pose estimation.
\textit{VIO only} also yields valid results in all conditions, but our method has
lower average ATE, suggesting that map observations improve accuracy by
suppressing drift.

\textit{Loosely coupled} and \textit{w/o pose feedback} fail to meet the criteria in some
conditions, suggesting that reevaluating correspondence geometry and
uncertainty through pose feedback helps maintain pose estimation.
Although \textit{Loosely coupled} yields valid results throughout ScanNet, its
higher average ATE than \textit{VIO only} may reflect bias from erroneous map
estimates.
Meanwhile, \textit{w/o recovery} yields valid results in all conditions but has
higher average ATE, which retrieval-based recovery may reduce by restoring
map constraints after occlusions or large camera rotations.

In covariance comparisons, both \textit{Fixed cov.} variants yield valid
results in all conditions on both datasets but have higher average ATE
than ours.
Additionally, \textit{Real-image head} has a higher average ATE than ours on
ScanNet and fails to meet the evaluation criteria in one in-house scene.
However, stable motion may provide sufficient VIO constraints on ScanNet,
limiting gains from covariance estimation.
Larger gains on the in-house dataset, with occlusions and large camera
rotations, suggest that weighting map correspondences by their uncertainty
helps suppress errors in challenging conditions.

\section{Conclusions}

We proposed SplatTether, which tightly couples 3DGS map-based localization
with VIO in a sliding-window factor graph to suppress drift and maintain
pose estimation when map correspondences are insufficient.
Learned covariances of keypoints on rendered images weight correspondence
factors, jointly estimated poses guide rendering, and retrieval supports recovery.
Experiments demonstrate the accuracy benefits of tight coupling and
learned correspondence uncertainty.

Future work will assess long-term stability of continuous pose estimation
and relocalization over larger areas with repeated transitions between
mapped and unmapped regions.

\clearpage

\section*{Acknowledgments}
This work was conducted by the National Institute of Advanced Industrial
Science and Technology (AIST) as part of FRONTia, the Multimodal Foundation
Model Development Project for AI Robots and Physical AI launched by the
Ministry of Economy, Trade and Industry (METI) in partnership with the New
Energy and Industrial Technology Development Organization (NEDO).

\section*{LLM Usage Statement}

We used OpenAI tools (ChatGPT/GPT and Codex) to edit and restructure text,
obtain reviewer-style feedback on manuscript sections, and assist with
related-work comparisons, citation organization, figures, data preparation,
and refinement of research ideas, claims, and conclusions.
For code implementation and experimental work, the authors reviewed,
edited, or approved tool-generated contributions.
The authors take full responsibility for all submitted content.

{
    \small
    \bibliographystyle{ieeenat_fullname}
    \bibliography{references}
}

\clearpage
\section*{\LARGE Supplementary Material}
\suppressfloats[t]

\section{Details of the Proposed Method}

\subsection{Algorithm Overview}

\begin{algorithm}[t]
  \caption{Online Visual-Inertial Localization}
  \label{alg:online-localization}
  \footnotesize
  \algrenewcommand\algorithmicrequire{\textbf{Input:}}
  \algrenewcommand\algorithmicensure{\textbf{Output:}}
  \begin{algorithmic}[1]
    \Require Image $\bm{I}_k^{\mathrm{obs}}$ and intervening IMU samples
    \Statex \textbf{Given:} 3DGS map, retrieval database,
      covariance head, and sensor calibration
    \Statex \textbf{State:} Sliding window, map factors, initialization
      buffer, retrieval request, and pending recovery pose
    \Ensure Updated state and a map-frame pose when available
    \State Advance VIO using image tracks, IMU, and registered map factors
    \If{an IMU bias norm exceeds its threshold}
      \State Reset VIO, map alignment, and matching state
      \State \Return
    \EndIf
    \State Choose retrieval if uninitialized, retrieval is requested,
      or recovery confirmation is pending. Otherwise choose local matching
    \State Set $\bm{T}_k^{\mathrm{rend}}$ to the retrieved or current camera pose
    \State $\mathcal{A}\gets\Call{MapAssociations}
      {\bm{I}_k^{\mathrm{obs}},\bm{T}_k^{\mathrm{rend}}}$
    \If{the map frame is not initialized}
      \State Buffer $\mathcal{A}$ and \Call{TryMapInitialization}{}
    \EndIf
    \If{the map frame is initialized}
      \If{local matching is used}
        \State Filter $\mathcal{A}$ by bearing consistency
        \State Register valid, uncertainty-weighted correspondence factors
        \State Request retrieval if bearing support is insufficient
      \Else
        \State $\hat{\bm{T}}_{C,k}\gets\Call{RansacPnP}{\mathcal{A}}$
        \If{\Call{ConfirmRecovery}{$\hat{\bm{T}}_{C,k},\mathcal{A}$}}
          \State Register a pose factor from $\hat{\bm{T}}_{C,k}$
        \EndIf
      \EndIf
    \EndIf
  \end{algorithmic}
\end{algorithm}

Algorithm~\ref{alg:online-localization} summarizes one backend update.
The 3DGS map, retrieval database, and covariance head are prepared offline.
The correspondence set $\mathcal{A}$ contains observed 2D keypoints,
their associated 3D map points, covariances, and validity information.

\medskip
\noindent\textbf{MapAssociations.}
This procedure renders a reference view, matches XFeat features with
LighterGlue, and lifts matches using rendered mean depth. It predicts
covariances of keypoints on the rendered image and propagates them to the
resulting map points.

\medskip
\noindent\textbf{TryMapInitialization.}
This procedure estimates and validates a shared gravity-constrained 4-DoF
alignment from buffered correspondences and a pending VIO candidate.
It commits the map-frame states only after registering initial correspondence
factors, then performs the first joint optimization.

\medskip
\noindent\textbf{ConfirmRecovery.}
This procedure validates PnP quality, the availability of both VIO states
in the active window, and pose information rank.
It retains the first valid candidate and forces retrieval
at the next attempt. It accepts only the second candidate when both belong
to the same estimation segment, meet the time interval bound, and yield
relative motion consistent with VIO,
then clears the pending candidate.
Invalid or inconsistent candidates also clear the pending candidate.

VIO updates optimize the active graph and marginalize states when applicable.
New map factors enter subsequent backend updates, except that successful
map initialization includes the first joint optimization.
We follow VINS-Mono~\cite{vinsmono2018} for feature tracking and sliding-window
settings.
Table~\ref{tab:supp-online-config} summarizes the main implementation settings.

\subsection{Local Matching}

We use at most 512 XFeat keypoints for local matching and set the LighterGlue
match threshold to 0.1.
We render a reference image from the jointly optimized camera pose and register
matches with the observed image as correspondence factors without running PnP.
The bearing error threshold increases from $1^\circ$ to $5^\circ$ in $1^\circ$
steps, stopping at the first threshold with at least 10 consistent
correspondences.
If fewer than 10 are found, we still register factors when at least one valid
correspondence remains and attempt image retrieval at the next matching frame.

\subsection{Map Frame Initialization and Recovery}

Before localization, we render the 3DGS map at the camera poses of the images
used for map optimization and build a NetVLAD retrieval database from the
renderings.
For map frame initialization, we accumulate correspondences across multiple
frames from the top three retrieved viewpoints.
We use them to estimate translation and yaw, the four degrees of freedom after
constraining the gravity direction.
We accept the estimate only if it has inliers from at least three frames, at
least 50 inliers in total, an inlier ratio of at least 20\%, and a reprojection
RMSE of at most $5\,\mathrm{px}$.
Before registering the initial map factors, we also check the Hessian rank and
condition number and gravity consistency.

After initialization, retrieval-based recovery requires at least 50 PnP
inliers, an inlier ratio of at least 50\%, and a reprojection RMSE of at most
$10\,\mathrm{px}$.
We retain the first candidate satisfying these criteria, the pose information
rank requirement, and active-state timing, then retrieve again at the next attempt.
We register a recovery pose factor only for the second candidate in the same
estimation segment if its timestamp and frame index increase, the time interval
is at most $1\,\mathrm{s}$, and the relative motions from retrieval and VIO
differ by at most $1\,\mathrm{m}$ in translation and $15^\circ$ in rotation.
We do not add correspondence factors from the same observation.

\subsection{Factor Construction and Optimization}

Correspondences from local matching are factor candidates if they lie within
the image and have valid covariances.
We propagate the rendered-image pixel covariance to 3D by backprojecting at a
fixed mean depth, then compute the predicted reprojection covariance for each
candidate.
If there are more than 128 candidates, we prioritize smaller determinants of
the predicted reprojection covariance and retain at most two points per
$0.05\,\mathrm{m}$ 3D voxel.
We then select up to 128 points in round-robin order across bins of image
location, depth, and viewing direction.

During optimization, we recompute the reprojection covariance at each
evaluated pose.
We use the Levenberg-Marquardt method and whiten the reprojection Jacobians by
this covariance during linearization.
Local correspondence factors contribute to the marginalization prior if the
prior passes validation.
Recovery pose factors remain active only while their states are in the sliding
window and are excluded from the marginalization prior.

\begin{table}[t]
  \centering
  \caption{Implementation settings.}
  \label{tab:supp-online-config}
  \small
  \begin{tabular}{@{}>{\raggedright\arraybackslash}p{.51\linewidth}>{\raggedright\arraybackslash}p{.41\linewidth}@{}}
    \toprule
    Parameter & Value \\
    \midrule[0.3pt]
    Frontend input rate (target) & $30\,\mathrm{Hz}$ \\
    Backend submission interval & Every 3 frontend frames \\
    Backend queue size & 2 frames \\
    VIO window size & 10 \\
    Accelerometer bias reset threshold & $2.5\,\mathrm{m/s^2}$ \\
    Gyroscope bias reset threshold & $1.0\,\mathrm{rad/s}$ \\
    \addlinespace[.3em]
    Map matching interval & Every backend frame \\
    Retrieval candidates (normal operation) & 1 viewpoint \\
    Local matching bearing threshold & $1^\circ$ to $5^\circ$ in $1^\circ$ steps \\
    Retrieval trigger at next matching frame & Fewer than 10 consistent local correspondences \\
    \addlinespace[.3em]
    Correspondence factor limit after initialization & 128 points \\
    3D voxel size and point limit & $0.05\,\mathrm{m}$, at most 2 points per voxel \\
    \addlinespace[.3em]
    VIO image factor Cauchy scale & 1 \\
    Reprojection covariance eigenvalue floor & $10^{-6}\,\mathrm{px}^2$ \\
    \bottomrule
  \end{tabular}
\end{table}

\begin{figure}[!t]
  \centering
  \includegraphics[width=\linewidth]{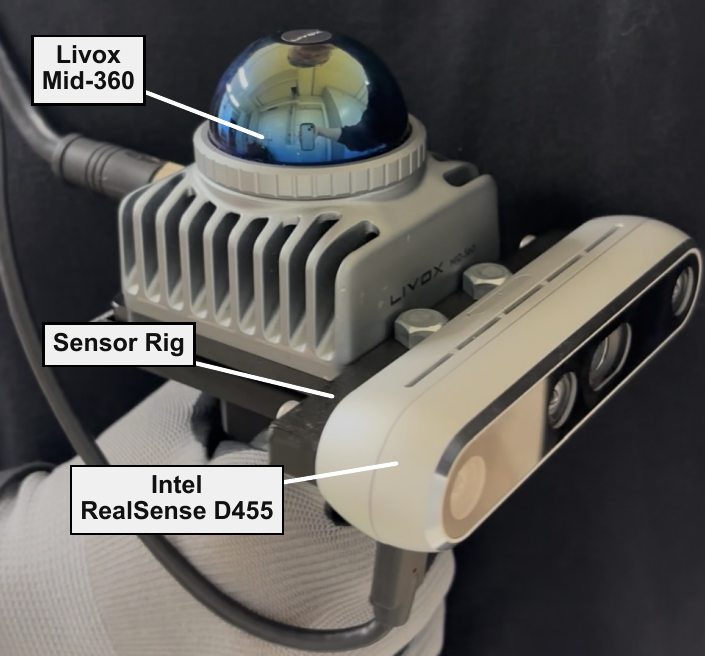}
  \caption{Sensor configuration used to collect the in-house dataset.}
  \label{fig:sensor-rig}
\end{figure}

\section{Experimental Details}

\subsection{In-House Dataset}

Figure~\ref{fig:sensor-rig} shows the sensor configuration used to collect the
in-house dataset.
A Livox Mid-360 and an Intel RealSense D455 are mounted on a rigid rig.
We aligned each LiDAR scan with a previously acquired point cloud of the
environment to obtain the ground-truth LiDAR trajectory.
We calibrated the LiDAR-to-camera extrinsics using separate calibration data
and the LiDAR-camera calibration toolbox~\cite{calib2023}.
We then transformed the LiDAR trajectory to obtain the ground-truth RealSense
camera trajectory.

\begin{figure*}[t]
  \centering
  \includegraphics[width=.99\textwidth]{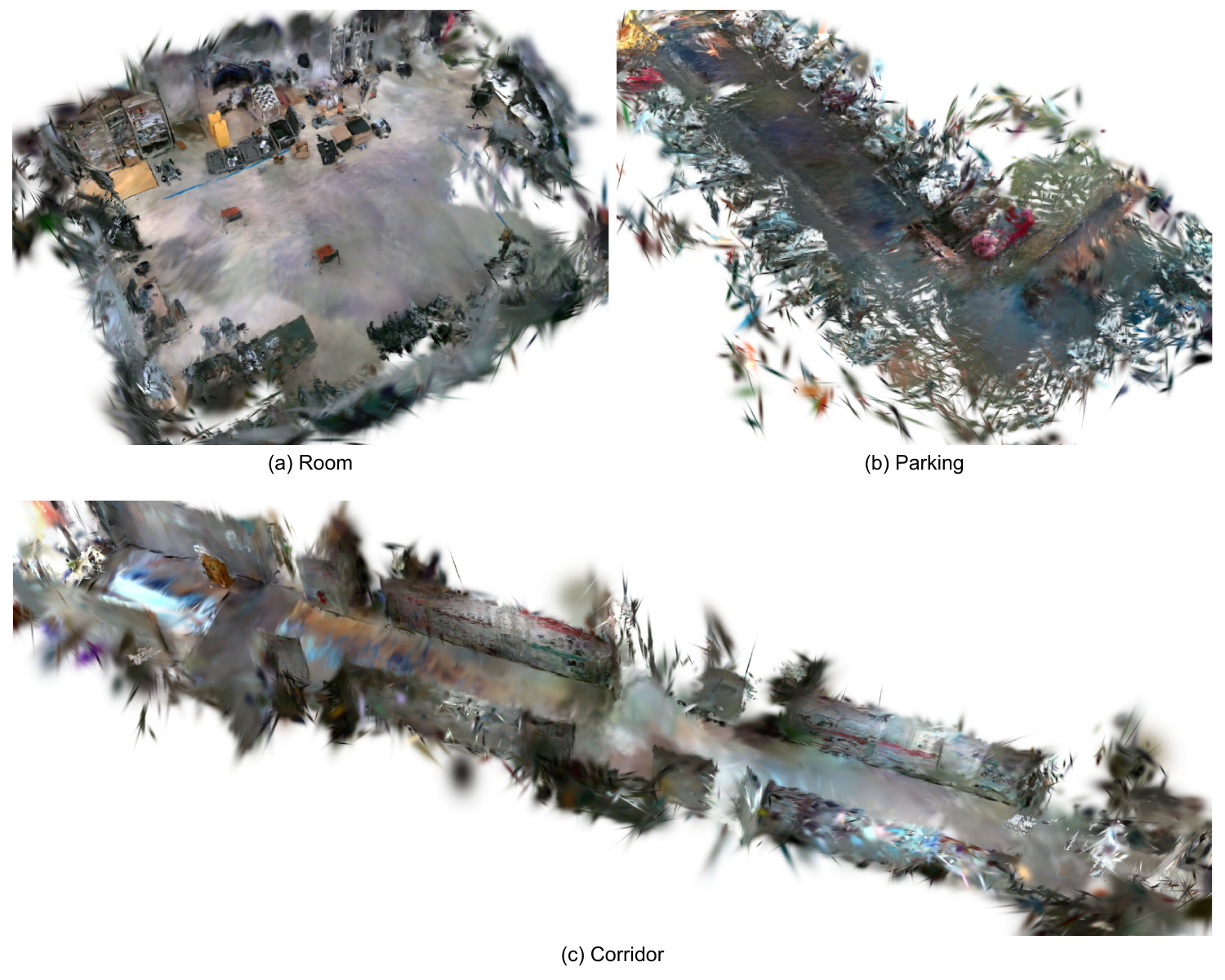}
  \caption{Evaluation 3DGS maps for the in-house dataset.}
  \label{fig:supp-inhouse-maps}
\end{figure*}

Figure~\ref{fig:supp-inhouse-maps} shows the 3DGS maps used for evaluation
in Room, Parking, and Corridor.

\subsection{ScanNet Dataset}

\begin{figure*}[p]
  \centering
  \includegraphics[width=.99\textwidth]{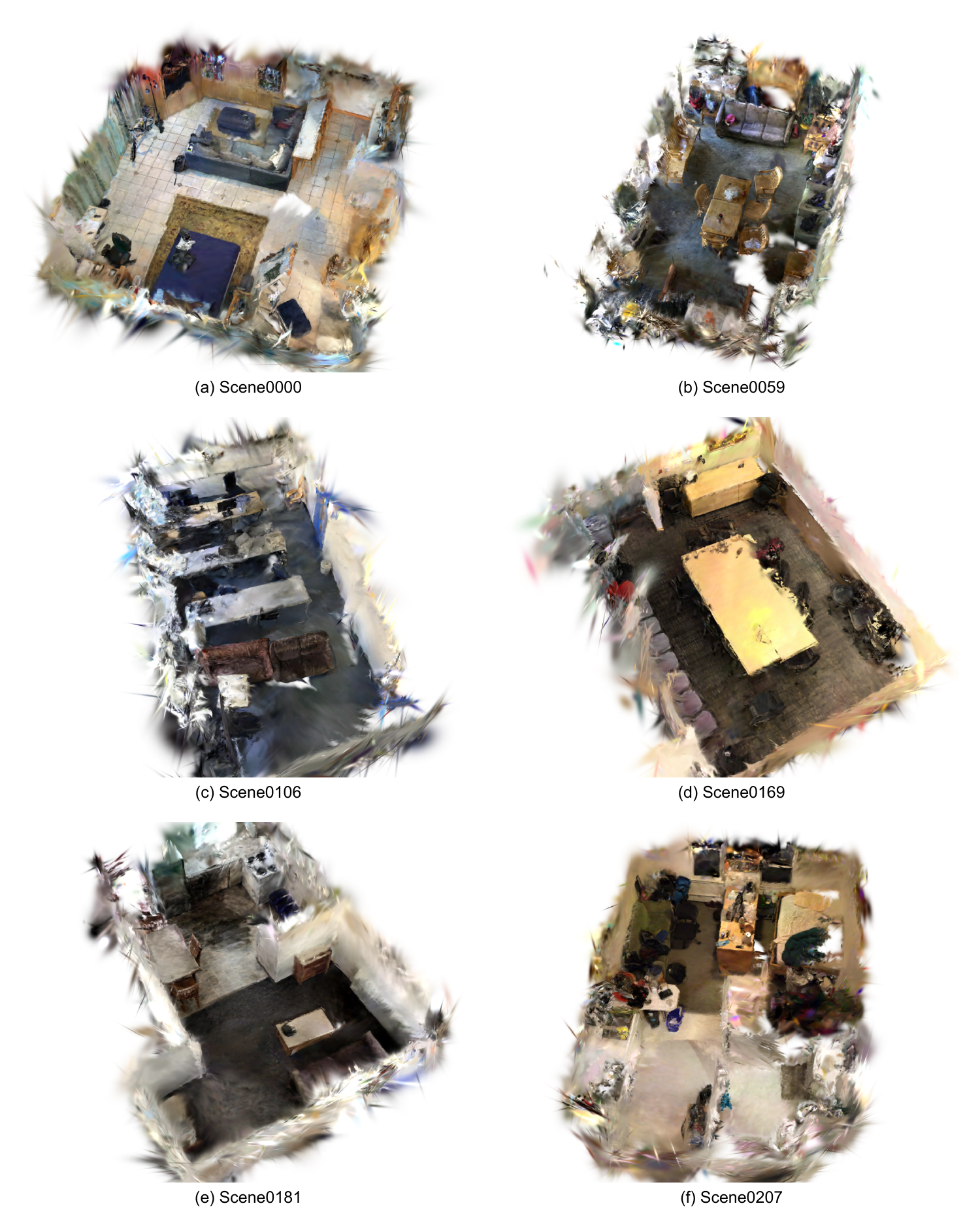}
  \caption{Evaluation 3DGS maps for six ScanNet scenes.}
  \label{fig:supp-evaluation-maps}
\end{figure*}

For each ScanNet scene, we use the \texttt{\_00} session to construct the map
and the \texttt{\_01} session for evaluation.
Figure~\ref{fig:supp-evaluation-maps} shows the 3DGS maps of the six ScanNet
scenes used for evaluation.

For localization evaluation, we register the \texttt{\_01} mesh to the
\texttt{\_00} mesh and apply the resulting fixed rigid transform to the
\texttt{\_01} ground-truth camera trajectory to obtain reference poses
in the map frame.

For our ScanNet experiments, we synthesize IMU measurements from the ground-truth
camera trajectory with the OpenVINS~\cite{openvins2020} simulator, treating the
camera and IMU frames as coincident.
The simulator interpolates the trajectory with an $\mathrm{SE}(3)$ B-spline, computes ideal
angular velocity and accelerometer readings while accounting for gravity, and
adds white Gaussian noise and random-walk biases.
Table~\ref{tab:supp-imu-noise} lists the parameters used for synthetic IMU generation.

\begin{table}[t]
  \centering
  \caption{Synthetic IMU generation parameters.
  Noise parameters are continuous-time coefficients shared by all three axes.}
  \label{tab:supp-imu-noise}
  \footnotesize
  \setlength{\tabcolsep}{3pt}
  \begin{tabular}{@{}lc@{}}
    \toprule
    Parameter & Value \\
    \midrule[0.3pt]
    IMU sampling rate [$\mathrm{Hz}$]
      & $200$ \\
    Gyro noise density [$\mathrm{rad}/(\mathrm{s}\sqrt{\mathrm{Hz}})$]
      & $1.6968\times10^{-4}$ \\
    Accel. noise density [$\mathrm{m}/(\mathrm{s}^{2}\sqrt{\mathrm{Hz}})$]
      & $2.0000\times10^{-3}$ \\
    Gyro bias random walk [$\mathrm{rad}/(\mathrm{s}^{2}\sqrt{\mathrm{Hz}})$]
      & $1.9393\times10^{-5}$ \\
    Accel. bias random walk [$\mathrm{m}/(\mathrm{s}^{3}\sqrt{\mathrm{Hz}})$]
      & $3.0000\times10^{-3}$ \\
    \bottomrule
  \end{tabular}
\end{table}

\subsection{3DGS Map Construction Settings}

Our 3DGS map construction pipeline is based on DN-Splatter~\cite{dnsplatter2025}.
We initialize the Gaussians from a point cloud generated using depth images and
camera poses from the mapping session, followed by voxel downsampling and normal
estimation.
We add and split Gaussians according to the loss gradients with respect to
their projected positions and remove Gaussians with low opacity.
For map optimization, we combine the RGB loss with an edge-aware log-L1 depth
loss, a sensor-depth normal loss, a normal TV loss, and a minimum-scale
loss.

Table~\ref{tab:supp-map-config} summarizes the settings for constructing the
evaluation maps.
The limit of $500{,}000$ Gaussians applies during initialization,
optimization, and final pruning.

\begin{table}[!ht]
  \centering
  \caption{Evaluation 3DGS map construction settings.}
  \label{tab:supp-map-config}
  \small
  \begin{tabular}{@{}>{\raggedright\arraybackslash}p{.51\linewidth}>{\raggedright\arraybackslash}p{.41\linewidth}@{}}
    \toprule
    Parameter & Value \\
    \midrule[0.3pt]
    Map inputs & RGB, depth, and mapping-session camera poses \\
    Validation frames & Every 10th frame of the source sequence \\
    Map training frames & Every 5th nonvalidation frame \\
    \addlinespace[.3em]
    Valid depth range & $0.1$ to $8.0\,\mathrm{m}$ \\
    Initial point voxel size & $0.01\,\mathrm{m}$ \\
    Gaussian limit & $500{,}000$ \\
    Spherical harmonics degree (maximum) & 3 \\
    \addlinespace[.3em]
    Optimization iterations & $30{,}000$ \\
    Position learning rate (initial, final) & $1.6\times10^{-4}$, $1.6\times10^{-6}$ \\
    Initial color feature learning rate & $2.5\times10^{-3}$ \\
    Initial opacity learning rate & $5.0\times10^{-2}$ \\
    Initial scale learning rate & $5.0\times10^{-3}$ \\
    Initial rotation learning rate & $1.0\times10^{-3}$ \\
    \addlinespace[.3em]
    RGB loss weights (L1, DSSIM) & $0.8$, $0.2$ \\
    Depth, normal, normal TV loss weights & $0.2$, $0.1$, $0.1$ \\
    Minimum-scale loss weight & $1.0$ \\
    \addlinespace[.3em]
    Densification schedule & Iterations 600 to 15,000, every 100 iterations \\
    \bottomrule
  \end{tabular}
\end{table}

\subsection{Covariance Head Training}

We train the rendered-image head using RGB-D frames excluded from 3DGS map
construction. To expose the head to variations in rendering quality, we train
it on maps constructed with depth and normal regularization, RGB-only
supervision, or RGB-only supervision with perturbed mapping poses.
For each frame, we independently choose maximum yaw and pitch
offsets from $5^\circ$ to $45^\circ$ in $5^\circ$ steps. We use zero, the
positive and negative half offsets, and the positive and negative full
offsets on each axis, producing 25 rendering viewpoints.
Figure~\ref{fig:rendering-covariance-training} shows nine such views.

We use a training match if the reprojection error of its map point in the
real image at the ground truth pose is at most $10\,\mathrm{px}$ and the
rendered mean depth differs from the sensor-derived depth
$d_i^{\mathrm{ref}}$ in the virtual camera frame by at most
$0.30\,\mathrm{m}+0.10d_i^{\mathrm{ref}}$.
The head receives a 64-dimensional rendered XFeat descriptor, its
64-dimensional absolute difference from the observed descriptor, five depth
geometry values, and LighterGlue match confidence. It maps the 133 feature
values to 32 dimensions, concatenates the match confidence, and uses
$33\to32\to3$ layers to produce the outputs used to construct the covariance
in Section~\ref{subsec:uncertainty-model}.
We set the numerical covariance floor to
$\delta_{\mathrm{cov}}=10^{-4}\,\mathrm{px}^2$.

\subsection{Online Estimation Settings}
\begin{table}[t]
  \centering
  \caption{IMU noise settings for state estimation.}
  \label{tab:supp-estimator-imu}
  \footnotesize
  \setlength{\tabcolsep}{2pt}
  \begin{tabular}{@{}lcc@{}}
    \toprule
    Parameter & In-house & ScanNet \\
    \midrule[0.3pt]
    Gyro noise density [$\mathrm{rad}/(\mathrm{s}\sqrt{\mathrm{Hz}})$]
      & $2.0544\times10^{-4}$ & $1.0\times10^{-3}$ \\
    \addlinespace[.3em]
    Accel. noise density [$\mathrm{m}/(\mathrm{s}^{2}\sqrt{\mathrm{Hz}})$]
      & $2.0765\times10^{-3}$ & $1.0\times10^{-2}$ \\
    \addlinespace[.3em]
    Gyro bias random walk [$\mathrm{rad}/(\mathrm{s}^{2}\sqrt{\mathrm{Hz}})$]
      & $1.1106\times10^{-5}$ & $1.0\times10^{-4}$ \\
    \addlinespace[.3em]
    Accel. bias random walk [$\mathrm{m}/(\mathrm{s}^{3}\sqrt{\mathrm{Hz}})$]
      & $4.1328\times10^{-4}$ & $1.0\times10^{-2}$ \\
    \bottomrule
  \end{tabular}
\end{table}

For the in-house dataset, we adopt the published D455 IMU noise settings
provided with OpenVINS~\cite{openvins2020}.\footnote{\url{https://github.com/rpng/open_vins/blob/master/config/rs_d455/kalibr_imu_chain.yaml}}
For ScanNet, we use conservatively inflated IMU noise coefficients relative to
those used for synthetic IMU generation.
Table~\ref{tab:supp-estimator-imu} summarizes the estimator noise settings.

\subsection{Detailed Ablation Results}
\label{sec:supp-ablation}

\begin{table}[t]
  \centering
  \caption{Translation ATE RMSE [m]$\downarrow$ for the component ablation
  on in-house data after 4-DoF alignment over the full trajectory
  (median of three runs).}
  \label{tab:supp-ablation-inhouse}
  \footnotesize
  \setlength{\tabcolsep}{2pt}
  \begin{tabular}{@{}lccccccc}
    \toprule
    & \multicolumn{2}{c}{Room} & \multicolumn{2}{c}{Corridor} & \multicolumn{2}{c}{Parking} & \multirow{2}{*}{Avg.} \\
    \cmidrule(lr){2-3}\cmidrule(lr){4-5}\cmidrule(lr){6-7}
    Method & Easy & Hard & Easy & Hard & Easy & Hard & \\
    \midrule[0.3pt]
    VIO only & 0.143 & 0.208 & 2.575 & 2.080 & 0.643 & 0.928 & \second{1.096} \\
    Map only & - & - & 3.253 & - & - & - & - \\
    Loosely coupled & 0.224 & 0.222 & \best{0.100} & 6.826 & - & - & - \\
    w/o pose feedback & 3.782 & 3.278 & 2.316 & 17.782 & - & - & - \\
    w/o recovery & \second{0.067} & \best{0.086} & 0.231 & 5.644 & \best{0.224} & 0.725 & 1.163 \\
    \addlinespace[2pt]
    \cdashline{1-8}[1.2pt/1.8pt]
    \addlinespace[2pt]
    Fixed cov. (default) & 0.071 & 0.093 & \second{0.154} & 7.133 & 0.378 & 10.772 & 3.100 \\
    Fixed cov. (train) & \best{0.065} & 0.088 & 0.227 & 5.755 & 0.244 & \best{0.564} & 1.157 \\
    Real-image head & \second{0.067} & 0.088 & 0.180 & \second{1.950} & \second{0.233} & - & - \\
    \textbf{Ours (full)} & \second{0.067} & \second{0.087} & 0.209 & \best{0.837} & 0.244 & \second{0.670} & \best{0.352} \\
    \bottomrule
  \end{tabular}
\end{table}

\begin{table}[t]
  \centering
  \caption{Translation ATE RMSE [m]$\downarrow$ for the component ablation
  on ScanNet after 4-DoF alignment over the full trajectory
  (median of three runs).}
  \label{tab:supp-ablation-scannet}
  \footnotesize
  \setlength{\tabcolsep}{2pt}
  \begin{tabular}{@{}lccccccc}
    \toprule
    Method & 0000 & 0059 & 0106 & 0169 & 0181 & 0207 & Avg. \\
    \midrule[0.3pt]
    VIO only & 0.230 & 0.175 & \best{0.099} & 0.207 & 0.180 & 0.115 & 0.168 \\
    Map only & 0.127 & 0.413 & 0.935 & 0.884 & 0.586 & 0.149 & 0.516 \\
    Loosely coupled & 0.052 & 0.101 & 0.114 & 0.820 & 0.107 & \best{0.083} & 0.213 \\
    w/o pose feedback & 0.057 & 0.143 & 0.712 & 1.312 & - & \second{0.084} & - \\
    w/o recovery & \best{0.051} & \best{0.090} & \second{0.106} & 0.113 & \second{0.082} & 0.086 & \second{0.088} \\
    \addlinespace[2pt]
    \cdashline{1-8}[1.2pt/1.8pt]
    \addlinespace[2pt]
    Fixed cov. (default) & 0.053 & 0.100 & 0.132 & 0.120 & 0.095 & 0.087 & 0.098 \\
    Fixed cov. (train) & \best{0.051} & 0.093 & 0.115 & 0.116 & \best{0.074} & 0.086 & 0.089 \\
    Real-image head & \best{0.051} & 0.097 & 0.131 & \best{0.097} & 0.084 & 0.087 & 0.091 \\
    \textbf{Ours (full)} & \best{0.051} & \best{0.090} & \second{0.106} & \second{0.104} & 0.083 & 0.086 & \best{0.087} \\
    \bottomrule
  \end{tabular}
\end{table}

Tables~\ref{tab:supp-ablation-inhouse} and~\ref{tab:supp-ablation-scannet}
report the per-sequence results underlying the dataset averages in
Table~\ref{tab:gsvins_ablation_results}.
For \textit{Fixed cov. (train)}, the standard deviation is the square root of the
mean squared pixel residual per coordinate, averaged first within each
training scene and then across scenes.

Using the median of three runs per condition, our method reduces average ATE
by approximately 48\% relative to \textit{VIO only} on ScanNet and lowers average
ATE on the in-house dataset, meeting the evaluation criteria in all in-house
conditions.
The per-sequence results show low ATE with constraints from local matching
and VIO in Room and ScanNet, possibly leaving limited room for further gains
from recovery or per-correspondence covariance estimation.
In Corridor Hard, however, our method achieves lower ATE than both
\textit{w/o recovery} and \textit{Fixed cov. (train)} in all three runs.
This suggests that retrieval-based recovery and per-correspondence covariance
estimation help limit error growth when local matching alone cannot
adequately control errors.

\end{document}